\documentclass[letterpaper, 10 pt, conference]{ieeeconf}  

\IEEEoverridecommandlockouts                              
\usepackage[bookmarks=true]{hyperref}
\usepackage{tabularx}
\usepackage{graphicx}
\usepackage{epsfig}
\usepackage{subfig}
\usepackage{cite}
\usepackage{float}
\usepackage{ifthen}
\usepackage{xcolor}
\usepackage{ntheorem}
\usepackage{xcolor}
\theoremseparator{:}
\usepackage{soul}
\usepackage{colortbl}
\usepackage{booktabs}
\let\labelindent\relax
\usepackage{enumitem}

\usepackage[font=footnotesize]{caption}
\setlist[itemize]{leftmargin=*}
\usepackage[flushleft]{threeparttable}
\usepackage{enumerate}
\usepackage{algorithm}
\usepackage{algpseudocode}
\algrenewcommand\algorithmicindent{1.2em}
\usepackage{amsmath,amssymb}

\usepackage{censor}

\definecolor{HElightgray}{RGB}{232, 232, 232}
\definecolor{Voicelightred}{RGB}{255, 173, 173}
\definecolor{E2lightblue}{RGB}{220, 234, 247}
\definecolor{E3lightblue}{RGB}{166, 202, 236}

\usepackage{stfloats}

\newif\ifanonymous
\anonymousfalse   

\ifanonymous
    
\else
    \StopCensoring
    
\fi

\newcommand{\fig}[1]{Fig.~\ref{#1}}

\title{\LARGE \bf Pack It My Way: Triadic Human-Robot Collaboration for Personalized Autonomous Packing}

\ifanonymous
\author{Anonymous Authors}
\else
\author{Sandeep Chowdary Kotapati$^*$, Yanxin Gao$^*$, and Tsung-Chi Lin%
\thanks{The authors are with the Department of Computer Science, New Jersey Institute of Technology, Newark 07102, NJ, USA (email: sk3898@njit.edu; yg446@njit.edu; tsungchi.lin@njit.edu).}
\thanks{$^*$Both authors contributed equally to this research.}
}
\fi

\begin{document}

\bstctlcite{IEEEexample:BSTcontrol}

\maketitle
\thispagestyle{empty}
\pagestyle{empty}

\begin{abstract}

Personalized autonomous packing requires robots to account for resident preferences that cannot be inferred from scene geometry alone. Expert teleoperators can interpret these preferences and translate them into feasible robot actions, but continuous expert involvement limits scalable deployment. In this paper, we investigate triadic human--robot collaboration among a resident, a correction mediator, and a robot by comparing human-expert and voice-agent mediation. We evaluate the two conditions in a user study across \textit{Protection}, \textit{Compactness}, and \textit{Grouping} tasks, using a \textit{Show--Correct--Generalize} process to assess preference correction and subsequent generalization after the surrounding objects are rearranged. Results show that voice-agent mediation achieves outcomes comparable to human-expert mediation in two of the three preference categories, despite receiving shorter and less detailed instructions. Both mediators are similarly easy to use, although the human expert is perceived as more reliable. These findings demonstrate the potential of voice agents to reduce expert involvement while identifying perceived reliability and preference generalization as remaining challenges.

\end{abstract}

\section{Introduction}\label{sec:intro}

General-purpose robots are emerging as promising platforms for assisting people in everyday home environments. Unlike task-specific systems, these robots can potentially support a range of household activities, including packing belongings, organizing objects, and handling routine domestic chores~\cite{wu2023tidybot}. However, deploying robots in homes remains fundamentally challenging because household tasks are open-ended, physically situated, and highly dependent on individual preferences. A viable near-term strategy is to involve expert remote operators who can perform tasks through teleoperation~\cite{nicol2022survey} while providing demonstrations and corrections that support future robot autonomy~\cite{spencer2022expert}. Yet domestic robot assistance is not simply a dyadic interaction between a teleoperator and a robot. The robot also operates alongside a resident who holds situated knowledge about the objects and the desired outcome. In a packing scenario, the resident may know which objects are fragile, which items belong together, and whether the arrangement should prioritize protection, compactness, or accessibility. The teleoperator, by contrast, may better understand what the robot can physically accomplish given its gripper, perception, workspace, and manipulation limits.

\begin{figure}[t]
    \centering
    \includegraphics[width=1\linewidth]{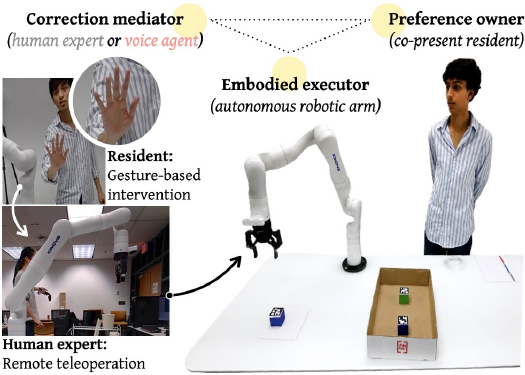}
    \caption{Overview of triadic human-robot collaboration for personalized autonomous packing. A co-present resident serves as the preference owner, a human expert or voice agent serves as the correction mediator, and an autonomous robotic arm serves as the embodied executor. The insets illustrate human-expert mediation, in which the resident provides gesture-based intervention and the remote expert teleoperates the robot.}
    \label{fig1}
    \vspace{-3ex}
\end{figure}

This setting creates a fundamentally \textit{triadic form of human-robot collaboration} involving a resident, a correction mediator, and a robot (\fig{fig1}). The resident specifies the desired outcome, the mediator translates that preference into an executable correction, and the robot performs and subsequently generalizes the corrected behavior. A human expert can provide this mediation during initial deployment by interpreting underspecified preferences and adapting them to the robot's capabilities. However, requiring continuous expert involvement limits the scalability of personalized assistance and the robot's progression toward greater autonomy. Recent advances in speech- and language-based robot interaction make a voice agent a promising alternative mediator: residents can express preferences through natural speech without directly controlling the robot, while the agent translates those preferences into robot actions~\cite{liu2023interactive,kuehn2026clarifying}. Nevertheless, spoken preferences may be incomplete or ambiguous, and existing language-based robot-learning methods primarily study direct dyadic correction rather than how an automated agent mediates between resident intent and robot execution. Prior multi-agent human-robot interaction research establishes the importance of interactions involving multiple human and robotic agents~\cite{gordon2023adaptive,dahiya2023survey}, but does not explain how replacing a human mediator with an automated one affects preference communication, perceived reliability, packing quality, and subsequent autonomous generalization.

We conducted a within-subjects user study with 12 participants to compare these two forms of mediation for personalized autonomous packing. In the \textit{human-expert-mediated condition}, the resident communicates their preference to an expert who interprets the guidance and teleoperates the robot. In the \textit{voice-agent-mediated condition}, the resident communicates the preference to an automated agent that translates it into a placement command for autonomous robot execution. We examined these conditions across three packing tasks derived from distinct categories of domestic preference: \textit{Protection}, representing safety- and constraint-related preferences; \textit{Compactness}, representing spatial-organization preferences; and \textit{Grouping}, representing semantic preferences. 

Each task follows a \textit{Show--Correct--Generalize} process. The robot first \textbf{shows} an initial placement, the resident \textbf{corrects} it through one of the mediators, and the robot then \textbf{generalizes} the acquired preference after the surrounding objects are rearranged. Results show that the voice agent achieved outcomes comparable to the human expert in two of the three preference categories, despite receiving shorter and less detailed instructions. 
%
%
Participants also found the two mediators similarly easy to use, although more participants considered the human expert more reliable. Overall, the findings demonstrate the potential of voice agents to reduce expert involvement in personalized autonomous packing while identifying preference generalization and perceived reliability as remaining challenges.

The main contributions of this paper are:

\noindent
(C1) \textit{Triadic human--robot collaboration} for allowing a resident to convey personalized packing preferences through either a human expert or a voice agent to an embodied robot;

\noindent
(C2) \textit{Show--Correct--Generalize personalized autonomous packing} for acquiring relational preferences and reapplying them after changes to the packing arrangement across Protection, Compactness, and Grouping tasks; and

\noindent
(C3) \textit{User study evaluation} for understanding whether voice-agent mediation can achieve outcomes comparable to human-expert mediation and generalize learned preferences to unseen packing configurations, while examining communication efficiency and user perceptions.

\section{Background and Related Work}\label{sec:related}

Personalized autonomous packing lies at the intersection of robotic manipulation, learning from human feedback, and multi-agent human--robot interaction. Prior work has developed autonomous systems for packing and object organization, methods for learning user preferences from demonstrations and corrections, and frameworks for collaboration involving more than two agents. In this section, we review these areas and identify the gaps addressed by this work.

\subsection{Autonomous Robotic Packing}
Autonomous packing requires a robot to determine both where objects should be placed and in what sequence, while satisfying geometric, stability, and manipulation constraints. Existing methods have primarily focused on warehouse and logistics settings, optimizing packing density while maintaining stable and executable placements~\cite{wang2022dense}. More recent work has incorporated human knowledge: Santos et al.\ learned packing sequences from human demonstrations~\cite{santos2024learning}, while iPack uses foundation models to infer semantic constraints, such as avoiding placements that could damage fragile items~\cite{blei2025ipack}. These approaches improve geometric feasibility and human-like packing behavior, but generally define the desired outcome through fixed objectives, aggregated demonstrations, or model-derived common-sense constraints. They provide limited support for correcting a placement according to one resident's situated preferences and applying that correction when the surrounding arrangement changes.

\subsection{Robot Learning from Human Preferences}
Human feedback enables robots to learn task objectives that cannot be inferred from the physical scene alone. Interactive learning methods allow people to refine robot behavior through corrective and evaluative feedback~\cite{chisari2022correct}, and unified frameworks combine demonstrations, corrections, and preference comparisons within a common reward-learning process~\cite{mehta2024unified}. Natural language provides a more accessible feedback channel: prior systems have incorporated online language corrections into shared-autonomy manipulation~\cite{cui2023right} and used verbal feedback to improve long-horizon robot policies~\cite{shi2024yell}. In domestic organization, TidyBot further demonstrates that robots can infer and generalize individual cleanup preferences from a small number of examples~\cite{wu2023tidybot}.

However, these studies primarily consider a direct interaction between a human teacher and a robot learner. They investigate whether a robot can learn from a particular type of feedback, but do not isolate how the agent mediating that feedback affects what the user communicates, how the correction is executed, or whether the resulting behavior is trusted and successfully generalized.

\subsection{Triadic Human-Robot Collaboration}
Multi-agent human-robot interaction (HRI) extends dyadic interaction to settings involving multiple humans, robots, or computational agents with different roles, communication channels, and control capabilities~\cite{dahiya2023survey}. Triadic frameworks have been developed for assistive tasks in which human and technological agents contribute complementary physical capabilities and objectives~\cite{gordon2023adaptive}. Recent robotic architectures have also supported two concurrent users by adapting object alignment to facilitate their joint physical interaction~\cite{semeraro2025trihrcbot}. Together, these studies show that adding a third agent changes how agency, knowledge, and control are distributed during collaboration.

Nevertheless, prior triadic systems largely assume a fixed team composition and prescribed role for each agent. Less attention has been given to \textit{preference mediation}, in which one person owns the task preference, a second human or automated agent translates that preference into an executable correction, and a robot performs and generalizes the corrected behavior. Consequently, it remains unclear whether an automated voice agent can occupy this mediator role with outcomes comparable to those achieved by a human expert.

\subsection{Knowledge and Research Gap}
Despite progress in autonomous packing, preference learning, and multi-agent HRI, three key gaps remain:

\noindent1) \textbf{Limited consideration of situated packing preferences} ---
Existing packing systems predominantly optimize predefined geometric, physical, or semantic objectives. They do not sufficiently address preferences that vary across residents and contexts, such as whether a placement should prioritize \textit{Protection}, \textit{Compactness}, or \textit{Grouping}.

\noindent2) \textbf{Lack of comparative evidence for preference mediation} ---
Prior learning-from-feedback research typically assumes that the user communicates directly with the robot through a single interaction channel. It remains unknown whether a voice agent can translate resident preferences into effective corrections with objective and subjective outcomes comparable to human-expert mediation.

\noindent3) \textbf{Missing evaluation of preference generalization} ---
Most studies evaluate whether feedback improves the current execution. Less is known about whether a corrected preference remains effective after the surrounding objects are rearranged and how residents perceive the resulting generalized behavior.

This paper addresses these gaps through a triadic collaboration framework comprising a resident, a correction mediator, and a robot. We compare \textit{human-expert-mediated} and \textit{voice-agent-mediated} correction across three categories of packing preference using a \textit{Show--Correct--Generalize} process. This comparison examines how mediator identity influences preference communication, correction quality, perceived reliability, satisfaction, and subsequent autonomous generalization.

\section{Triadic Human-Robot Collaboration}\label{sec:proposed}

We formulate personalized autonomous packing as a triadic collaboration involving three functionally distinct roles. The \textit{preference owner} determines what constitutes an acceptable outcome based on contextual and personal knowledge. The \textit{correction mediator} interprets the owner's preference and translates it into a correction that is feasible for the robot. The \textit{embodied executor} perceives and manipulates the physical environment while retaining and subsequently generalizing the corrected preference. Roles are therefore distinguished by their knowledge and responsibilities rather than by whether the agent is human or computational. In our implementation, a co-present resident serves as the preference owner, a robotic manipulator serves as the embodied executor, and either a remote human expert or a voice agent serves as the correction mediator.

\subsection{Preference Owner}

The resident serves as the preference owner because they possess situated knowledge unavailable to either the mediator or the robot. This knowledge includes how belongings should be treated, which outcomes are personally acceptable, and which contextual relationships should be preserved. The resident's objective is not to specify robot trajectories or low-level manipulation commands, but to communicate the desired packing outcome after observing the robot's behavior.

Our preliminary exploration of resident-guided packing identified five broad categories of preference: (1) \textit{safety and constraint preferences}, including fragility sensitivity, weight distribution, and collision tolerance; (2) \textit{spatial-organization preferences}, including packing density, layout structure, and accessibility; (3) \textit{semantic preferences}, including grouping by object type, owner, or intended use; (4) \textit{aesthetic and subjective preferences}, including neatness, symmetry, and personal habits; and (5) \textit{privacy and boundary constraints}, including restricted objects and areas the robot should avoid. These preferences may affect either skill-level execution, such as speed or approach direction, or task-state decisions, such as object selection and placement. This paper narrows this broader space to safety and constraint, spatial organization, and semantic preferences. These categories allow distinct relational placement preferences to be operationalized and objectively evaluated while holding the objects, robot platform, and overall packing activity constant. 

\subsection{Embodied Executor}

\begin{figure}[t]
    \centering
    \includegraphics[width=1\linewidth]{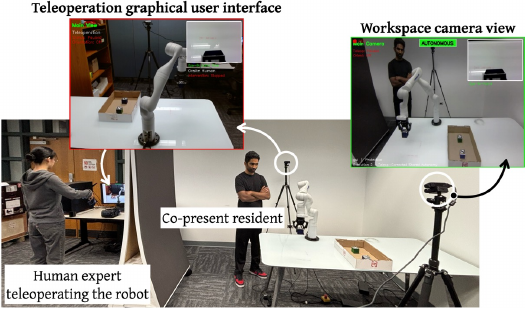}
    \caption{Camera views and graphical interfaces supporting human-expert teleoperation and autonomous robot execution.}
    \label{fig2}
    \vspace{-3ex}
\end{figure}

The embodied executor is a 7-DoF Kinova Gen3 robotic arm equipped with a Robotiq 2F-85 parallel-jaw gripper. A stationary Orbbec Gemini 2 camera provides a global view of the workspace and supports the detection of object-mounted ArUco markers~\cite{garrido2014automatic} and the resident's hand gestures. A second camera mounted on the end effector supports visual servoing during object pickup and close-range marker localization. The global-camera, robot-base, and packing-container coordinate frames are jointly calibrated. A webcam positioned behind the robot provides the remote human expert with a live view of the robot and packing workspace. The human-expert teleoperation GUI presents this webcam feed as its main view and the end-effector camera feed as an inset, while a separate global workspace view supports monitoring during autonomous execution (see \fig{fig2}).

\begin{figure*}[t]
    \begin{center}
    \vbox{ 
    \includegraphics[width=\textwidth]{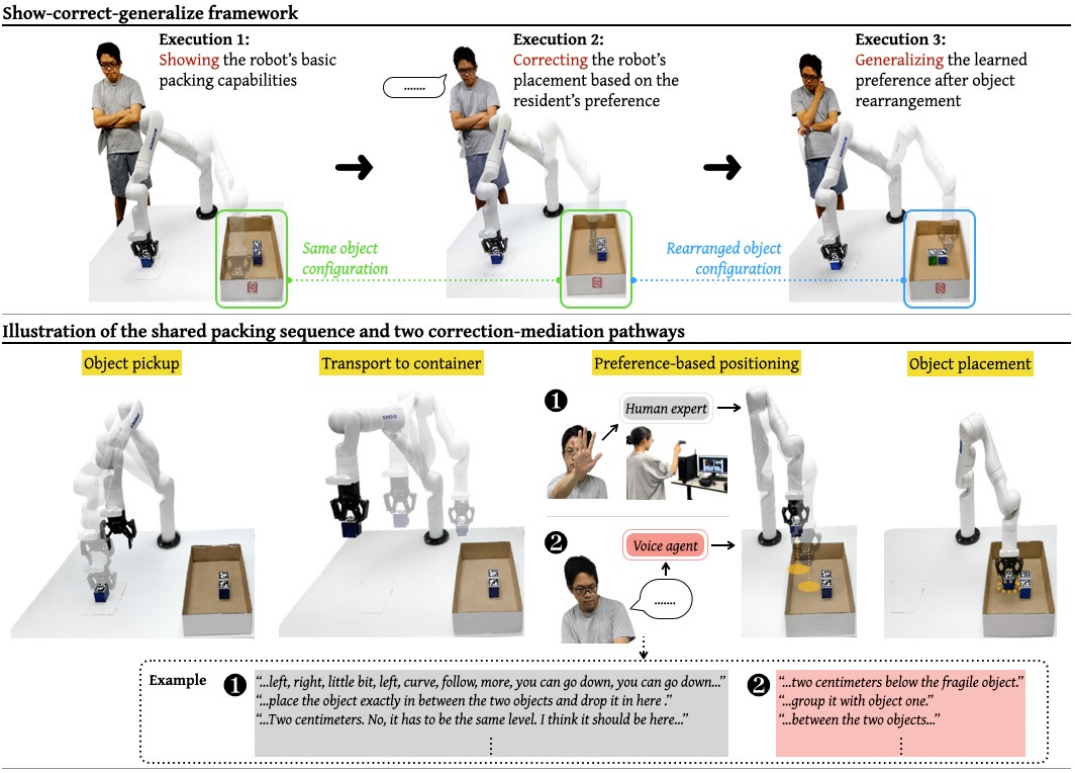} 
    }
    \caption{Show--Correct--Generalize framework and the two correction-mediation pathways. \textbf{Top:} The robot first demonstrates its basic autonomous packing behavior (Execution 1), receives a resident-specific placement correction under the same object configuration (Execution 2), and generalizes the learned relational preference after the reference objects are rearranged (Execution 3). \textbf{Bottom:} Both pathways share autonomous object pickup and transport. During preference-based positioning, the resident provides correction through either (1) a remote human expert who teleoperates the robot following an open-palm intervention or (2) a voice agent that converts a spoken instruction into a placement target. Representative correction inputs are shown at the bottom.}
    \label{fig3}
    \vspace{-3ex}
    \end{center}
\end{figure*}

The robot autonomously executes basic pick-and-place behaviors through a perception-and-control pipeline that estimates the target object's pose in the robot-base frame and transports it to a specified target using Cartesian-space pose control. This basic autonomy determines how the robot perceives, grasps, and transports the object, but not where the resident wants it placed. Preference learning is therefore handled separately by updating the relational placement goal while retaining the same underlying manipulation pipeline.

\subsection{Correction Mediator}

The correction mediator bridges the resident's contextual knowledge and the robot's physical execution capabilities. We select two realizations that represent a progression from expert-supported robot learning toward more scalable automated mediation (\fig{fig3}, bottom). Human-expert mediation reflects an established approach in which an expert teleoperates the robot while providing demonstrations and corrections~\cite{spencer2022expert}. Voice-agent mediation offers a natural interaction channel for novice residents and can reduce the need for continuous expert involvement by translating spoken preferences directly into robot actions~\cite{liu2023interactive,kuehn2026clarifying}.

\noindent
\colorbox{HElightgray}{\textbf{Human-Expert-Mediated Correction:}}
In this approach, a remote expert observes the workspace through a graphical user interface displaying the webcam feed. The robot begins its autonomous placement behavior, and the resident may request an intervention using an open-palm gesture (\fig{fig1}), at which point the robot stops and control transfers to the expert. The expert teleoperates the arm using an HTC VIVE handheld controller, which maps the expert's natural hand motions to the robot end effector. The interface incorporates the RelaxedIK solver~\cite{rakita2017motion,boguslavskii2023shared} to generate smooth and feasible robot motions while avoiding joint-space discontinuities, kinematic singularities, and self-collisions. This motion-tracking approach has been shown to provide more intuitive robot teleoperation than alternative interfaces~\cite{lin2022intuitive}, making it well suited for providing real-time corrections during robot execution.

During teleoperation, the resident provides continuous verbal guidance, while the expert interprets this guidance in relation to the robot's reachability, gripper geometry, and workspace constraints. When the resident is satisfied, the expert releases the object. The resulting placement is recorded as the resident-specific correction.

\noindent
\colorbox{Voicelightred}{\textbf{Voice-Agent-Mediated Correction:}}
In this condition, the resident communicates the desired correction through a single spoken instruction. Audio is captured using a US-OB-33 microphone, resampled to 16~kHz, and transcribed locally using the English-only Whisper \texttt{small.en} model through the \texttt{faster-whisper} implementation~\cite{radford2023robust}. The resulting transcript is processed using a predefined keyword-to-action mapping, without employing a general-purpose large language model (LLM) for semantic interpretation or response generation. Because the correction vocabulary is constrained and each keyword maps directly to a predefined robot action, an LLM is unnecessary. Excluding an LLM also reduces latency and computational overhead, keeps speech processing local, and avoids introducing variability from open-ended language reasoning, thereby isolating the effect of speech-mediated correction.

A deterministic parser interprets each transcript using the active preference category and converts it into a structured relational command containing a reference object, spatial relation, and optional distance. Numerical distances may be expressed in different units, while qualitative terms such as \textit{tightly} and \textit{loosely} are mapped to predefined distances. The resulting relation is converted into a Cartesian placement target, which the robot executes through its autonomous manipulation pipeline. Table~\ref{tab:voice-command} summarizes the supported instruction structures and examples. Because the mapping is rule-based, the same instruction and scene state produce the same placement target, enabling an interpretable and repeatable comparison.

\begin{table*}[t]
\caption{Supported voice-command structures and examples across the three resident preference categories. Basic commands express a qualitative relation, whereas parameterized commands add a reference object, direction, and/or metric distance.}
\label{tab:voice-command}
\centering
\small
\renewcommand{\arraystretch}{1.18}
\setlength{\tabcolsep}{5pt}
\begin{tabularx}{\textwidth}{
    @{}
    p{0.2\textwidth}
    p{0.38\textwidth}
    X
    @{}
}
\toprule
\textbf{Preference Category} &
\textbf{Supported Instruction} &
\textbf{Example Voice Command(s)} \\
\midrule

Safety and constraint &
\textbf{Basic:} Relative safety constraint. 
\newline
\textbf{Parameterized:} Reference object or property with a direction and/or distance. &
\textit{``Between the two objects''; 
``3 cm from the fragile object''; 
``5 cm to the right of Object 2.''} \\

\midrule

Spatial organization &
\textbf{Basic:} Qualitative packing density.
\newline
\textbf{Parameterized:} Reference object with a direction and distance. &
\textit{``Pack them tightly''; 
``Pack them loosely''; 
``5 cm to the left of Object 1.''} \\

\midrule

Semantic &
\textbf{Basic:} Grouping relation with a reference object.
\newline
\textbf{Parameterized:} Reference object with a direction and distance. &
\textit{``Group it with Object 2''; 
``3 cm to the right of Object 1.''} \\

\bottomrule
\end{tabularx}
\vspace{-1ex}
\end{table*}

\noindent
\textbf{Updating Robot Autonomy:}
Both mediation pathways produce the same form of preference record: a preference category, a reference object or object configuration, and a direction--distance relationship. In the human-expert-mediated condition, this relationship is derived from the final teleoperated placement. In the voice-agent-mediated condition, it is generated directly from the parsed instruction.

The correction does not modify the robot's basic autonomous manipulation pipeline during the study. Instead, it updates the resident-specific relational placement goal supplied to the Cartesian-space pose controller. When the surrounding objects are rearranged, the system applies the stored relationship to their updated locations to compute a new placement target, which the robot executes using the same object-pickup and transport pipeline. This shared representation allows generalization quality to be compared across the two mediators while holding the underlying robot autonomy constant.

\section{User Study}\label{sec:exp}

We conducted a within-subjects user study to examine how human-expert and voice-agent mediation affect preference communication, packing quality, perceived reliability, and subsequent autonomous generalization. 

\subsection{Participants}

We recruited 12 participants (8 men and 4 women) from a local university to assume the resident role. Participants ranged in age from 19 to 40 years (\textit{M} = 24.1, \textit{SD} = 5.7) and reported moderate prior experience with robots (\textit{M} = 3.2, \textit{SD} = 1.1) on a five-point scale, with 5 indicating extensive experience. Each study session lasted approximately 90 minutes, and participants received \$15 in compensation.

\begin{figure}[b]
    \centering
    \vspace{-1ex}
    \includegraphics[width=1\linewidth]{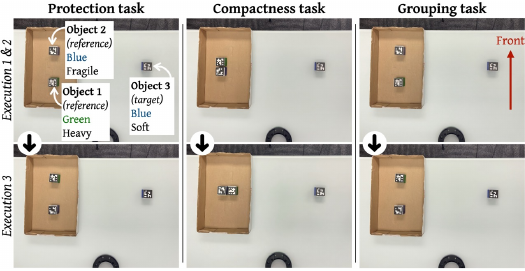}
    \caption{Object configurations across the three executions for the protection, compactness, and grouping tasks. Executions~1 and~2 use the same reference-object configuration (top). Before Execution~3 (bottom), reference Objects~1 and~2 are rearranged to create a previously unseen configuration for evaluating preference generalization. Object~3 is the target object in all executions. The red arrow indicates the front direction.}
    \label{fig4}
\end{figure} 

\subsection{Preference-Based Packing Tasks}

We designed three structured packing tasks to isolate distinct resident preference spaces: safety and constraint, spatial organization, and semantic organization. In each task, participants specified where Object~3 should be placed relative to two reference objects already inside the container (see \fig{fig4}). The objects shared the same geometry and mass but were distinguished by color and ArUco markers, allowing the preference context to vary while holding manipulation difficulty constant.

\begin{itemize}
    \item \textbf{Protection task} (\textit{safety and constraint}): Object~1 (green) was described as heavy, Object~2 (blue) as fragile, and Object~3 (blue) as soft. Participants specified how the soft target object should be placed to protect the fragile object from the heavy object.

    \item \textbf{Compactness task} (\textit{spatial organization}): Participants specified how tightly or loosely Object~3 should be packed relative to the two reference objects.

    \item \textbf{Grouping task} (\textit{semantic organization}): Participants specified that Object~3 should be placed near the reference object with which it shared an assigned semantic category.
\end{itemize}

\subsection{Show--Correct--Generalize Framework}

To evaluate whether each correction mediator could support learning a resident-specific preference from a single intervention and generalizing it beyond the corrected scene, each task followed a three-execution Show--Correct--Generalize sequence (\fig{fig3}, top):

\begin{itemize}
    \item \textbf{Show (execution~1)} --- \textit{establish a preference-agnostic baseline:} Without receiving preference information, the robot autonomously placed target Object~3 at a predefined location. This allowed the resident to observe the robot's initial behavior and identify the need for correction.

    \item \textbf{Correct (execution~2)} --- \textit{capture the resident's preference:} Under the same object configuration, the resident corrected the placement through either the human expert or the voice agent. The correction was encoded as a relational placement goal defined with respect to Objects~1 and~2.

    \item \textbf{Generalize (execution~3)} --- \textit{evaluate relational preference transfer:} Reference Objects~1 and~2 were rearranged into a previously unseen configuration. The robot applied the stored relationship to their updated positions and autonomously placed Object~3, testing whether it could generalize the preference rather than reproduce an absolute Cartesian position.
\end{itemize}

\subsection{Experimental Procedure}

After providing informed consent, participants completed a pre-study questionnaire covering demographics and prior experience with robots and voice assistants. They were then introduced to the robotic system, the three packing tasks, and both correction-mediation conditions. The experiment followed a within-subjects design comparing two correction mediators---a human expert and a voice agent---across three packing tasks. Because the \textit{Show} execution was independent of the mediator, each participant completed it once per task, followed by separate \textit{Correct} and \textit{Generalize} executions under each mediator.

In the voice-agent-mediated condition, participants received a reference card listing the accepted basic and parameterized command formats (Table~\ref{tab:voice-command}). In the human-expert-mediated condition, participants were not constrained by a command template and could provide continuous verbal guidance while the expert teleoperated the robot. The same human expert, who had more than 200 hours of robot teleoperation experience, conducted all expert-mediated trials to reduce operator-related variability.

The packing-task order was randomized for each participant. Before each execution, the scene was manually reset to the corresponding starting configuration. After each \textit{Correct} execution, participants evaluated the correction process. During each \textit{Generalize} execution, they observed the robot's autonomous placement and subsequently evaluated the generalization outcome. A final questionnaire elicited an overall comparison of the two mediators.

\subsection{Measures and Analyses}

\noindent\textbf{Correction Communication:}
We measured instruction word count and classified instruction complexity as \textit{Brief} (no more than 10 words and one relation or action), \textit{Moderate} (11--25 words, a quantified distance, or two to three actions), or \textit{Detailed} (more than 25 words, repeated refinements, or at least four actions). Trials without a spoken component were excluded only from the language analysis.

\noindent\textbf{Packing Quality:}
Let $\mathbf{p}_i$ denote the center position of Object~$i$. We computed a task-specific quality score in $[0,1]$:

\begin{equation}
\begin{aligned}
q_{\mathrm{P}} &=
\exp\left(
-\frac{d_{\perp}
(\mathbf{p}_3,\overline{\mathbf{p}_1\mathbf{p}_2})}
{\tau_{\mathrm{P}}}
\right),\\
q_{\mathrm{C}} &=
\exp\left(
-\frac{A(\mathbf{p}_1,\mathbf{p}_2,\mathbf{p}_3)}
{\tau_{\mathrm{C}}}
\right),\\
q_{\mathrm{G}} &=
1-\frac{d(\mathbf{p}_3,\mathbf{p}_{r})}
{d(\mathbf{p}_3,\mathbf{p}_{r})+
 d(\mathbf{p}_3,\mathbf{p}_{u})},
\end{aligned}
\label{eq:packing-quality}
\end{equation}

where $d_{\perp}$ is the perpendicular deviation from the line connecting Objects~1 and~2, $A$ is the triangle area formed by the three object centers, and $\mathbf{p}_{r}$ and $\mathbf{p}_{u}$ denote the related and unrelated objects, respectively. We set $\tau_{\mathrm{P}}$ to half the object width and $\tau_{\mathrm{C}}$ to the object footprint area, anchoring both scores to the physical object dimensions. Higher values indicate stronger satisfaction of the corresponding task criterion.


\noindent\textbf{Subjective Measures:}
Following the \textit{Correct} executions, participants selected which mediator made it easier to express their preferences and which they perceived as more reliable. After each \textit{Generalize} execution, participants rated their satisfaction with the robot's learning on a five-point scale.

\noindent\textbf{Analysis:}
For each packing task, we compared packing quality between the human-expert-mediated and voice-agent-mediated conditions using paired-samples TOST with equivalence bounds of $\pm0.10$ on the normalized 0--1 scale. This bound was informed by pilot robot executions and expert assessment of normal placement variability. Other outcome measures were compared using paired-samples $t$-tests when the within-participant differences were approximately normally distributed and Wilcoxon signed-rank tests otherwise. We applied Holm correction across the three task-specific comparisons for each outcome measure and execution stage. We also compared packing quality between \textit{Correct} and \textit{Generalize} within each mediator and task using the same distribution-appropriate paired tests. Finally, we used a linear mixed-effects regression model to examine the association between correction word count and packing quality, with word count, correction mediator, and packing task as fixed effects and participant as a random intercept.

\section{Results and Discussion}\label{sec:res}

This section examines how the correction mediator affected preference communication, subsequent generalization, and participant perceptions.

\begin{figure*}[!b]
    \begin{center}
    \vbox{ 
    \includegraphics[width=\textwidth]{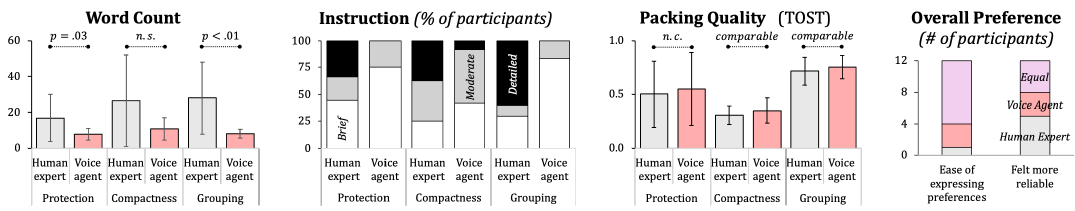} 
    }
    \caption{Correction communication, packing quality, and overall mediator preferences. The first three panels show \textit{Correct} (E2) results across packing tasks: word count, instruction complexity, and packing quality. The final panel summarizes overall preferences for ease of expression and reliability. \textit{n.s.} denotes a nonsignificant difference; \textit{comparable} and \textit{n.c.} indicate that equivalence was demonstrated or not demonstrated, respectively.}
    \label{fig5}
    \vspace{-3ex}
    \end{center}
\end{figure*}

\subsection{Correction Communication}

Participants used significantly fewer words with the voice agent than with the human expert for Protection ($p=.03$) and Grouping ($p<.01$). The same tendency appeared for Compactness, although the difference was not significant. Voice-agent instructions were also predominantly \textit{Brief} and rarely \textit{Detailed}, whereas human-expert-mediated corrections contained larger proportions of \textit{Moderate} and \textit{Detailed} instructions, particularly for Grouping. The mixed-effects regression found no significant overall association between correction word count and packing quality. Despite this reduced communication, paired TOSTs at the \textit{Correct} stage supported equivalent packing quality between the two mediation conditions for Compactness and Grouping (both $p<.05$), whereas equivalence was not demonstrated for Protection (see \fig{fig5}).

\textit{Toward scalable preference mediation ---}
The shorter voice-agent instructions and equivalent correction-stage packing quality for Compactness and Grouping suggest that voice agents can reduce continuous human-expert involvement for structured preferences supported by the command space. However, equivalence was not demonstrated for Protection, indicating that this benefit may not extend uniformly across preference types. Because additional verbal detail was not significantly associated with higher packing quality, human experts may instead serve as exception handlers for ambiguous, unsupported, or higher-stakes preferences, enabling more scalable personalized assistance and greater robot autonomy.

\subsection{Task-Dependent Preference Generalization}

\fig{fig6} shows that packing quality did not differ significantly between \textit{Correct} and \textit{Generalize} for Protection under either mediator. For Compactness, quality was significantly higher during \textit{Generalize} under both human-expert ($p<.01$) and voice-agent ($p<.05$) mediation. For Grouping, quality showed no statistically significant change under human-expert mediation but decreased significantly under voice-agent mediation ($p<.05$). Because no additional correction occurred before \textit{Generalize}, the increased Compactness scores do not indicate further learning; rather, the stored relational goal produced a higher-quality arrangement after the reference objects were repositioned.

\textit{Representation-dependent generalization ---}
The similar patterns observed for Protection and Compactness suggest that both mediators can support transferring geometric preferences expressed through constraints, directions, and distances. The decline in voice-agent Grouping performance highlights a limitation of converting semantic intent into a predefined spatial relationship. Although an instruction such as \textit{``group it with Object~2''} identifies the relevant reference object, a fixed direction--distance representation may not fully preserve the underlying category-level preference after the scene changes. Future voice mediators should retain semantic relationships explicitly and support clarification or target confirmation before execution.

\subsection{Ease of Expression and Perceived Reliability}

\begin{figure*}[t]
    \begin{center}
    \vbox{ 
    \includegraphics[width=\textwidth]{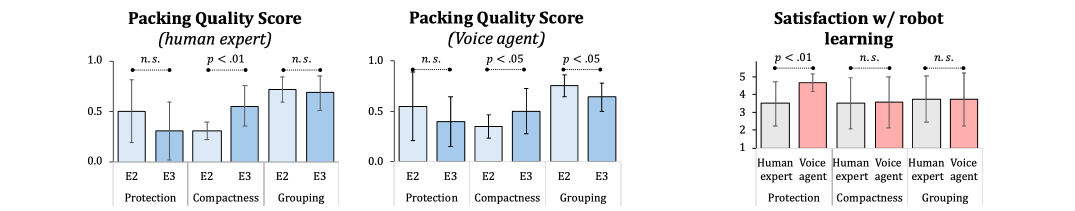} 
    }
    \caption{Packing-quality scores compare \textit{Correct} (E2) and \textit{Generalize} (E3) within each mediator and task, while the right panel compares satisfaction between mediators after \textit{Generalize}. Higher scores indicate better outcomes; \textit{n.s.} denotes a nonsignificant difference.}
    \label{fig6}
    \vspace{-4ex}
    \end{center}
\end{figure*}

Eight participants considered the two mediators equally easy for expressing preferences, while three selected the voice agent and one selected the human expert. Perceived reliability showed a different pattern: five participants selected the human expert, three selected the voice agent, and four considered them equal (\fig{fig5}, right). Participants also reported significantly greater satisfaction with the robot's learning under voice-agent mediation for Protection ($p<.01$), with no significant mediator differences for the other two (\fig{fig6}, right).

\textit{Scalable mediation requires calibrated trust ---}
Although the voice agent offered an accessible interaction channel, ease of expression did not necessarily translate into perceived reliability. Human experts may be viewed as better able to interpret ambiguity, provide continuous feedback, and recover from incorrect guidance. Voice-agent interfaces should therefore make their interpretations visible, confirm the inferred relational goal, and allow residents to revise or clarify a correction before robot execution.

\section{Conclusion}\label{sec:con}

We presented a triadic human--robot collaboration framework for personalized autonomous packing and used a Show--Correct--Generalize process to compare human-expert and voice-agent mediation. Voice-agent mediation elicited shorter and less detailed instructions, showed no significant decline in Protection quality, and yielded higher Compactness scores after rearrangement. However, Grouping quality declined, suggesting that predefined spatial representations do not fully preserve semantic preferences. Most participants rated the mediators equally easy to use, although more considered the human expert more reliable. These findings demonstrate the potential of voice agents to reduce continuous expert involvement when preferences can be expressed through structured relational commands; however, reliable automated mediation requires mechanisms that preserve semantic intent and communicate the agent's interpretation to the resident.

\textit{Limitations ---}
The study involved a small university sample, simplified objects, marker-based perception, a constrained voice-command vocabulary, and a single human expert. It also examined one-shot corrections rather than preferences learned through repeated household interactions. Future work should evaluate more diverse users, objects, experts, and packing contexts; support clarification and iterative correction; and investigate how preferences evolve through longer-term interaction.

\section{Acknowledgments}

\noindent
\textbf{Author CRediT:} Conceptualization (all); Data Curation (SCK, YG); Formal Analysis (all); Investigation (SCK, YG); Methodology (all); Software (SCK, YG); Supervision (TL); Visualization (TL); Original Draft (SCK, YG); Review \& Editing (TL).
\textbf{AI Statement:} Text edited with an LLM and verified for accuracy by the authors.

\bibliographystyle{IEEEtran}
\bibliography{references}

\begin{thebibliography}{10}
\providecommand{\url}[1]{#1}
\csname url@samestyle\endcsname
\providecommand{\newblock}{\relax}
\providecommand{\bibinfo}[2]{#2}
\providecommand{\BIBentrySTDinterwordspacing}{\spaceskip=0pt\relax}
\providecommand{\BIBentryALTinterwordstretchfactor}{4}
\providecommand{\BIBentryALTinterwordspacing}{\spaceskip=\fontdimen2\font plus
\BIBentryALTinterwordstretchfactor\fontdimen3\font minus \fontdimen4\font\relax}
\providecommand{\BIBforeignlanguage}[2]{{%
\expandafter\ifx\csname l@#1\endcsname\relax
\typeout{** WARNING: IEEEtran.bst: No hyphenation pattern has been}%
\typeout{** loaded for the language `#1'. Using the pattern for}%
\typeout{** the default language instead.}%
\else
\language=\csname l@#1\endcsname
\fi
#2}}
\providecommand{\BIBdecl}{\relax}
\BIBdecl

\bibitem{wu2023tidybot}
J.~Wu, R.~Antonova, A.~Kan, M.~Lepert, A.~Zeng, S.~Song, J.~Bohg, S.~Rusinkiewicz, and T.~Funkhouser, ``{TidyBot}: Personalized robot assistance with large language models,'' \emph{Autonomous Robots}, vol.~47, no.~8, pp. 1087--1102, 2023.

\bibitem{nicol2022survey}
M.~Nicol, L.~Lu, and C.~Wang, ``A survey study on the technology and public acceptance of remote labor,'' \emph{IFAC-PapersOnLine}, vol.~55, no.~27, pp. 416--423, 2022.

\bibitem{spencer2022expert}
J.~C. Spencer, S.~Choudhury, M.~Barnes, M.~Schmittle, M.~Chiang, P.~J. Ramadge, and S.~S. Srinivasa, ``Expert intervention learning: An online framework for robot learning from explicit and implicit human feedback,'' \emph{Autonomous Robots}, vol.~46, no.~1, pp. 99--113, 2022.

\bibitem{liu2023interactive}
H.~Liu, A.~Chen, Y.~Zhu, A.~Swaminathan, A.~Kolobov, and C.-A. Cheng, ``Interactive robot learning from verbal correction,'' \emph{arXiv preprint arXiv:2310.17555}, 2023.

\bibitem{kuehn2026clarifying}
H.~Kuehn, L.~Santos, and I.~Leite, ``Clarifying constraints in interactive robot learning with language feedback,'' in \emph{Proceedings of the 2026 ACM/IEEE International Conference on Human-Robot Interaction}.\hskip 1em plus 0.5em minus 0.4em\relax ACM, 2026, pp. 816--824.

\bibitem{gordon2023adaptive}
D.~F.~N. Gordon, A.~Christou, T.~Stouraitis, M.~Gienger, and S.~Vijayakumar, ``Adaptive assistive robotics: A framework for triadic collaboration between humans and robots,'' \emph{Royal Society Open Science}, vol.~10, no.~6, p. 221617, 2023.

\bibitem{dahiya2023survey}
A.~Dahiya, A.~M. Aroyo, K.~Dautenhahn, and S.~L. Smith, ``A survey of multi-agent human--robot interaction systems,'' \emph{Robotics and Autonomous Systems}, vol. 161, p. 104335, 2023.

\bibitem{wang2022dense}
F.~Wang and K.~Hauser, ``Dense robotic packing of irregular and novel 3d objects,'' \emph{IEEE Transactions on Robotics}, vol.~38, no.~2, pp. 1160--1173, 2022.

\bibitem{santos2024learning}
A.~Santos, N.~F. Duarte, A.~Dehban, and J.~Santos-Victor, ``Learning the sequence of packing irregular objects from human demonstrations: Towards autonomous packing robots,'' in \emph{2024 10th IEEE RAS/EMBS International Conference for Biomedical Robotics and Biomechatronics (BioRob)}, 2024, pp. 951--957.

\bibitem{blei2025ipack}
\BIBentryALTinterwordspacing
Y.~Blei, M.~Krawez, A.~G{\"o}{\ss}, D.~V. Sheela, T.~J{\"u}lg, P.~Krack, F.~Walter, and W.~Burgard, ``{iPack}: Intuitive bin packing with large language models,'' \emph{arXiv preprint arXiv:2503.08445}, 2025. [Online]. Available: \url{https://arxiv.org/abs/2503.08445}
\BIBentrySTDinterwordspacing

\bibitem{chisari2022correct}
E.~Chisari, T.~Welschehold, J.~Boedecker, W.~Burgard, and A.~Valada, ``Correct me if i am wrong: Interactive learning for robotic manipulation,'' \emph{IEEE Robotics and Automation Letters}, vol.~7, no.~2, pp. 3695--3702, 2022.

\bibitem{mehta2024unified}
S.~A. Mehta and D.~P. Losey, ``Unified learning from demonstrations, corrections, and preferences during physical human--robot interaction,'' \emph{ACM Transactions on Human-Robot Interaction}, vol.~13, no.~3, pp. 1--25, 2024.

\bibitem{cui2023right}
Y.~Cui, S.~Karamcheti, R.~Palleti, N.~Shivakumar, P.~Liang, and D.~Sadigh, ``No, to the right: Online language corrections for robotic manipulation via shared autonomy,'' in \emph{Proceedings of the 2023 ACM/IEEE International Conference on Human-Robot Interaction}, 2023, pp. 93--101.

\bibitem{shi2024yell}
L.~X. Shi, Z.~Hu, T.~Z. Zhao, A.~Sharma, K.~Pertsch, J.~Luo, S.~Levine, and C.~Finn, ``Yell at your robot: Improving on-the-fly from language corrections,'' in \emph{Proceedings of Robotics: Science and Systems}, Delft, Netherlands, Jul. 2024.

\bibitem{semeraro2025trihrcbot}
F.~Semeraro, J.~Leadbetter, and A.~Cangelosi, ``{TriHRCBot}: A robotic architecture for triadic human--robot collaboration through mediated object alignment,'' in \emph{2025 IEEE International Conference on Robotics and Automation (ICRA)}, 2025, pp. 6703--6709.

\bibitem{garrido2014automatic}
S.~Garrido-Jurado, R.~Mu{\~n}oz-Salinas, F.~J. Madrid-Cuevas, and M.~J. Mar{\'i}n-Jim{\'e}nez, ``Automatic generation and detection of highly reliable fiducial markers under occlusion,'' \emph{Pattern Recognition}, vol.~47, no.~6, pp. 2280--2292, 2014.

\bibitem{rakita2017motion}
D.~Rakita, B.~Mutlu, and M.~Gleicher, ``A motion retargeting method for effective mimicry-based teleoperation of robot arms,'' in \emph{Proceedings of the 2017 ACM/IEEE International Conference on Human-Robot Interaction}, 2017, pp. 361--370.

\bibitem{boguslavskii2023shared}
N.~Boguslavskii, Z.~Zhong, L.~M. Genua, and Z.~Li, ``A shared autonomous nursing robot assistant with dynamic workspace for versatile mobile manipulation,'' in \emph{2023 IEEE/RSJ International Conference on Intelligent Robots and Systems (IROS)}.\hskip 1em plus 0.5em minus 0.4em\relax IEEE, 2023, pp. 7040--7045.

\bibitem{lin2022intuitive}
T.-C. Lin, A.~U. Krishnan, and Z.~Li, ``Intuitive, efficient and ergonomic tele-nursing robot interfaces: Design evaluation and evolution,'' \emph{ACM Transactions on Human-Robot Interaction (THRI)}, vol.~11, no.~3, pp. 1--41, 2022.

\bibitem{radford2023robust}
A.~Radford, J.~W. Kim, T.~Xu, G.~Brockman, C.~McLeavey, and I.~Sutskever, ``Robust speech recognition via large-scale weak supervision,'' in \emph{Proceedings of the 40th International Conference on Machine Learning}, ser. Proceedings of Machine Learning Research, vol. 202.\hskip 1em plus 0.5em minus 0.4em\relax PMLR, 2023, pp. 28\,492--28\,518.

\end{thebibliography}

\end{document}